\documentclass[letterpaper, 10 pt, conference]{ieeeconf}
\IEEEoverridecommandlockouts
\usepackage[utf8]{inputenc}
\usepackage{graphicx}
\usepackage{multicol}
\usepackage{multirow}
\usepackage{array}
\usepackage{footmisc}
\usepackage{amssymb}
\usepackage{amsmath}
\usepackage{amsfonts}
\usepackage{algorithm}
\let\labelindent\relax
\usepackage[inline]{enumitem}
\usepackage{algpseudocode}
\usepackage{tikz}
\usepackage{float}
\usepackage{soul}
\usetikzlibrary{math,shapes}
\usepackage{tabularx}
\usepackage{pbox}
\usepackage{ragged2e}
\usepackage{balance}
\usepackage{subfig}
\usepackage{relsize}
\usepackage{tabu}
\usepackage{booktabs} %
\usepackage{multirow}
\usepackage{multicol}
\usepackage{adjustbox}

\usepackage{hyperref}

\usepackage{listings}

\mathchardef\mhyphen="2D

\newcommand{\TaskName}{Gasket Assembly}

\usepackage{etoolbox}
\makeatletter
\patchcmd{\@makecaption}
  {\scshape}
  {}
  {}
  {}
\makeatletter
\patchcmd{\@makecaption}
  {\\}
  { -\ }
  {}
  {}
\makeatother
\usepackage{siunitx}
\usepackage[backend=biber,
            hyperref=true,
            url=true,
            isbn=false,
            doi=false,
            backref=false,
            style=ieee,
            natbib=true,%
            mincitenames=1,
            maxcitenames=1,
            citestyle=numeric-comp,
            sorting=none,
            block=none,
            maxbibnames=99]{biblatex}
\usepackage{hyperref}

\title{\LARGE \bf Automating Deformable Gasket Assembly}
\author{Simeon Adebola$^{*1}$,  Tara Sadjadpour$^{*1}$,  Karim El-Refai$^{*1}$, Will Panitch$^1$, Zehan Ma$^1$, \\ Roy Lin$^1$,  Tianshuang Qiu$^1$, Shreya Ganti$^1$, Charlotte Le $^1$, Jaimyn Drake $^1$, and Ken Goldberg$^1$
\thanks{$^{*}$Equal Contribution}
\thanks{$^1$The AUTOLab at UC Berkeley (automation.berkeley.edu)
{\tt\small \{simeon.adebola, goldberg\} @berkeley.edu}}
}

\begin{document}
\maketitle
\thispagestyle{empty}
\pagestyle{empty}

\begin{abstract}
In \TaskName{}, a deformable gasket must be aligned and pressed into a narrow channel. This task is common for sealing surfaces in the manufacturing of automobiles, appliances, electronics, and other products. \TaskName{} is a long-horizon, high-precision task and the gasket must align with the channel and be fully pressed in to achieve a secure fit. To compare approaches, we present 4 methods for \TaskName{}: one policy from deep imitation learning and three procedural algorithms. We evaluate these methods with 100 physical trials. Results suggest that the Binary+ algorithm succeeds in 10/10 on the straight channel whereas the learned policy based on 250 human teleoperated demonstrations succeeds in 8/10 trials and is significantly slower.
Code, CAD models, videos, and data can be found at \href{https://berkeleyautomation.github.io/robot-gasket/}{https://berkeleyautomation.github.io/robot-gasket/}.
\end{abstract}

\section{Introduction}
\label{sec:intro}
Tasks such as clothes folding, thread untangling, and cable tracing have applications in manufacturing, logistics, and domestic applications, but present unique challenges for robots due to the complex physics and large configuration spaces of deformable objects. One such deformable manipulation task is the assembly of \textit{gaskets}, deformable components that fill the space between two or more mating surfaces to provide a seal, generally to prevent leakage from or into the joined objects while under compression~\cite{aibada2017review}. Gaskets compensate for small gaps or imperfections in mating surfaces and play critical roles in industries such as automotive and appliance manufacturing (where they are essential for sealing windows, engines, and fuel systems), plumbing, power generation, and construction. Almost all Gasket Assembly today is performed by humans.

We propose a robot gasket assembly task where a robot must pick and place the gasket, a 1D deformable object, into a channel of similar length and width, such that the gasket snugly and completely fills the entire channel. The setup can be easily and reliably replicated using common materials and a 3D printer with the provided CAD model. 
The task requires pick-and-place, press, and slide moves to be performed in succession. This is a high-precision, long-horizon task with a large state space, complicated dynamics, and low error tolerances\cite{viswanath_autonomously_2022}. 

\begin{figure}[ht!]
    \vspace{0.2cm}
    \centering
    \includegraphics[width=\linewidth]{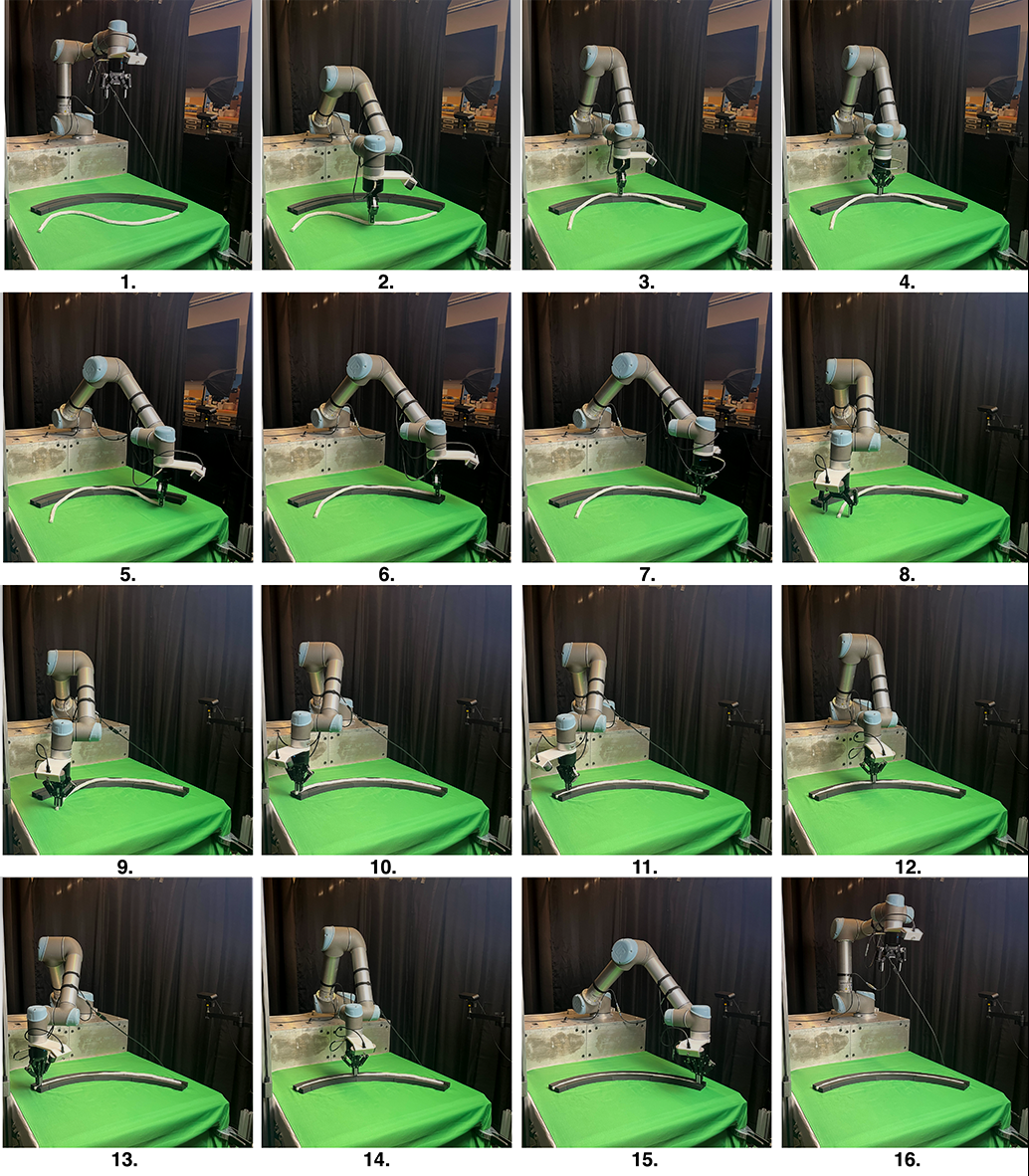}
    \caption{\textbf{Gasket Assembly Example.} This is an example of the Binary+ procedural algorithm rollout on the curved channel. Initially, the gasket is separate from the channel (1). The robot picks the gasket's midpoint (2), places it into the channel's midpoint (3), and presses down to insert (4). Then an endpoint of the gasket is picked, placed, and pressed into the endpoint of the channel (5,6,7); the same is done with the second gasket endpoint (8,9,10,11). We omit the remaining pick-place-press steps at the quarter and eighth points, as well as the second press that occurs at each point for reinforcing insertion. Finally, the gripper returns to the middle of the channel (12) and slides from the middle to one end (13). It returns to the middle and slides to the other endpoint (14,15). The end result is a successfully assembled gasket (16).}
    \vspace{-0.6cm}
\label{fig:r2d2}
\end{figure}

\TaskName{} qualifies as a dull/dreary repetitive task \cite{hagele2016} and is thus a candidate for robotic automation with the goal of reducing worker fatigue and cost as a result of the automation. We propose four approaches to automating the gasket assembly task: one learned implementation and three procedural. The procedural approaches differ primarily in the strategy that each uses to choose and order the pick-and place and press motions on the gasket, while the learned end-to-end approach utilizes Diffusion Policy~\cite{chi2023diffusionpolicy} to learn a gasket insertion policy from 250 human demonstrations.

This paper makes the following contributions:
\begin{enumerate}
    \item A novel formulation of a gasket assembly task that is easy to replicate.
    \item A deep imitation learning policy generated from  250 human-teleoperated demonstrations. 
    \item Three analytic/procedural algorithms for the same task.  
    \item Experimental results comparing the learned policy and the procedural algorithms based on 100 physical trials that suggests procedural algorithms can be superior in terms of performance. 
\end{enumerate}

\section{Related Work}
\label{sec:related}
\subsection{Deformable Linear Object Perception and Manipulation}
Deformable objects can be categorized into three distinct types based on their dimensions: 1D objects (e.g., ropes, cables, strings, and electrical cords~\cite{viswanath_autonomously_2022, viswanath2023ltodo, sundaresan_learning_2020, suberkrub_feel_2022}), 2D objects (e.g., fabrics and clothing~\cite{avigal_speedfolding_2022}), and 3D objects (e.g., bags~\cite{seita_learning_2021, chen_autobag_2022, chen_efficiently_2023}). In this paper, we focus on the 1D class of deformable objects, also known as deformable linear objects (DLOs), since they best represent gaskets. We use the terms ``DLO" and ``gasket" interchangeably in this paper. 

There have been recent advancements in robotics for a variety of manipulation tasks involving DLOs, such as knot tying~\cite{sundaresan_learning_2020}, rope untangling~\cite{viswanath_autonomously_2022}, and the production of wire harnesses~\cite{suberkrub_feel_2022}. Nonetheless, DLO perception and manipulation remain open research areas. In this paper, we explore the assembly of gaskets, expanding upon the current scope of robotic manipulation for DLOs.

The typical approach for handling DLOs involves a three-step process: state estimation, path planning, and manipulation to achieve the desired outcome~\cite{kabul2007cable, keipour_efficient_2022}. Achieving the target state of a DLO can be complex, necessitating path planning strategies that can adapt to and recover from sub-optimal actions. A wide range of methods have been explored to solve different aspects of this problem, including Coherent Point Drift for DLO state estimation~\cite{8403315}, learning from demonstrations for DLO manipulation~\cite{schulman_learning_2016, 6943185, 7759418, chen2023fast, schrum2023reciprocal}, and self-supervised learning for understanding the state-space of DLOs~\cite{yan_self-supervised_2020}. In this work, we present 4 baselines that encompass a variety of methods seen in these prior works. The analytical algorithms use a computer vision algorithm to detect the  channel and gasket, and primitives to help the robot recover from sub-optimal actions taken in prior steps. The learned baseline learns from human demonstrations using diffusion policy~\cite{chi2023diffusionpolicy}.

\subsection{Cable Routing Tasks with DLOs}
A common task category for 1D deformable manipulation is routing, in which a DLO is manipulated to match a given set of position and shape constraints.

Recent research has explored a range of techniques and approaches for ``cable routing," a process in which a cable is guided along a path by a series of unconnected fixtures ~\cite{jin_robotic, waltersson_planning_2022, monguzzi_tactile, wilson_cable, luo2023multistage}. Some studies have focused specifically on the path planning aspect of routing, aiming to navigate the cable from its starting position to a designated endpoint~\cite{keipour_efficient_2022}. Another rendition of this task involves guiding a string or rope through openings that are only slightly larger than the material itself, with a minimal clearance of 1.4mm~\cite{7139532}. In contrast with these scenarios that involve threading a rope through a series of openings or guiding cables along a path, this paper focuses on the precise challenge of snugly fitting a gasket into a specifically shaped channel. This task demands a greater level of accuracy in both perception and manipulation, highlighting the need for advanced techniques capable of handling the intricacies of secure gasket insertion.

\subsection{Insertion Tasks}
Robot manipulation has been applied to a number of insertion tasks involving both rigid and non-rigid objects. The most classic example is the peg-hole insertion task. Insertion tasks have been widely explored using learning from demonstrations~\cite{wang_demo, tang2015learning, tang2016teach}, reinforcement learning~\cite{schoettler_meta, zhao_off, zhang_learning, schoettler_deep},  regression\cite{spector_ins} and multimodal perception~\cite{lee2020making}, where the peg and insertion hole are rigid. Another work on insertion~\cite{fu_safely_2022} uses visual and tactile sensors and force-torque sensing together with self-supervised learning to achieve policies that allow a robot to insert a USB connector in an industrial task setting. ~\cite{luo2018deep} uses reinforcement learning techniques to learn how to insert a rigid peg into a deformable hole. Peg-hole insertion has recently been applied to medical settings in~\cite{adagolodjo2019robotic}, which studies techniques to insert a flexible needle into a deformable foam. These iterations of the peg-hole insertion task differ from gasket assembly which requires the gasket to deform into the rigid channel.

Pirozzi and Natale~\cite{8458233} focuses on wire insertion, where a robot gripper informed by signals from its tactile sensor inserts a wire into a hole as part of a switchgear assembly.

\subsection{Robot Policy Learning for Long-Horizon Tasks}
In long-horizon sequential manipulation tasks, earlier actions affect the feasibility of subsequent actions~\cite{pirk2020modeling, driess_learning}. Recent approaches include the use of imitation and reinforcement learning~\cite{lynch2019play, schmalstieg_learn,wang_learning}, language-conditioned policy learning~\cite{mees2022calvin}, self-supervised learning~\cite{nair2019hierarchical} and learning from human demonstrations\cite{mandlekar2021learning}. \TaskName{} is a long-horizon task with subtasks that include picking, placing, and insertion.

\subsection{Deep Imitation Learning and Diffusion Policy}
Deep imitation learning from human teleoperated demonstrations is an active area of research. Diffusion Policy~\cite{chi2023diffusionpolicy}, motivated by the powerful generative modeling capabilities of diffusion models~\cite{sohl-dickstein_deep, ho_denoising}, was recently proposed to represent a robot’s visuomotor policy as a conditional denoising diffusion process. It learns the gradient of the action-distribution score function during training and iteratively performs a series of stochastic Langevin dynamics steps during inference. Specifically, starting from  $\mathbf{a}^K$ sampled from Gaussian noise, the Denoising Diffusion Probabilistic Model (DDPM)  performs K iterations of denoising to produce intermediate vectors with decreasing levels of noise, $\mathbf{a}^K, \mathbf{a}^{K-1}, ..., \mathbf{a}^0$, until a desired level of noise is obtained. Mathematically, to learn the conditional distribution $p(\mathbf{a}_t | \mathbf{o}_t)$, where $\mathbf{o}_t$ is the observation of the current step and $\mathbf{a}_t$ is the desired action output, we use a conditional CNN $\mathbf{\epsilon}_\theta(\mathbf{a}^k, k | \mathbf{o})$ to get 
$$\mathbf{a}^{k-1} = \alpha (\mathbf{a}^k - \gamma \mathbf{\epsilon}_\theta(\mathbf{a}^k, k, \mathbf{o}) + \mathcal{N}(0, \sigma^2 I)),$$
where $\alpha, \gamma, \sigma$ are noise schedule hyperparameters and functions of the iteration step $k$. During training, we minimize
$$\mathcal{L} = MSE(\mathbf{\epsilon}^k, \mathbf{\epsilon}_\theta(\mathbf{a}^k, k, \mathbf{o})),$$
where $\mathbf{\epsilon}^k$ is a random noise with appropriate variance.

\section{The Gasket Assembly Problem}
\label{sec:cfp}

We propose a problem where an automated system must reliably insert a deformable gasket into a rigid channel of predefined shape and length. Both the channel and the gasket are continuous. In this paper, we consider gaskets with circular cross-section.  We propose two human evaluation metrics: (1) Alignment: How well does the gasket align with the target shape of the channel and (2) Insertion: How much of the gasket is contained within the channel.

\begin{figure}
    \vspace{0.2cm}
    \centering
    \includegraphics[scale=0.3]{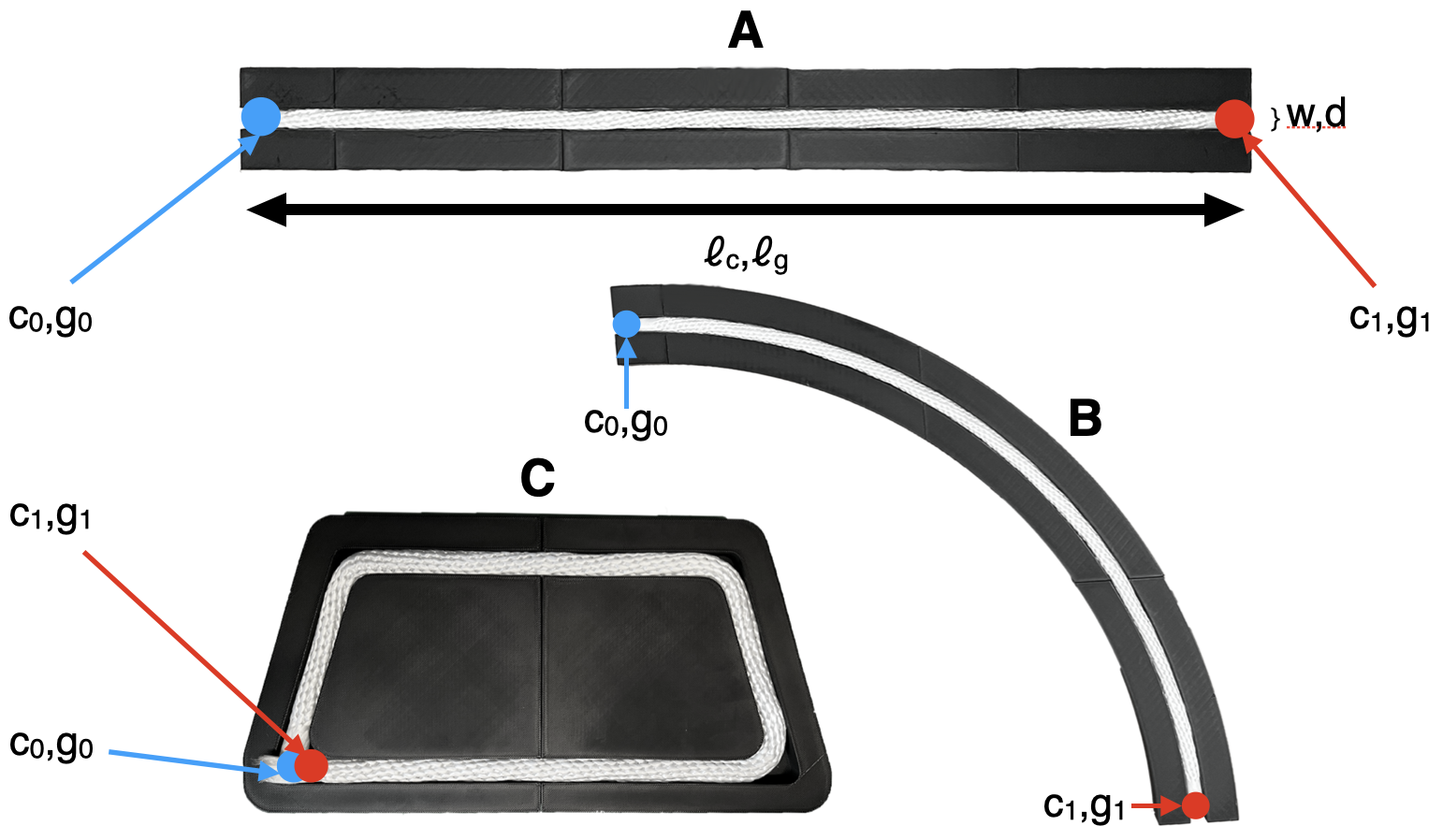}
    \caption{\textbf{Channels and Gaskets in Goal Positions. }Each channel has a gasket fully inserted. The straight channel (A) and the curved channel (B) are both open-ended channels whereas the trapezoid channel (C) is closed. This means that for all channels, the gasket endpoints ($g_0, g_1$) and channel endpoints ($c_0, c_1$) lie nearly on top of each other, but in the trapezoid case, $c_0$ and $c_1$ also lie nearly on top of each other.}    \vspace{-0.5cm}
\label{fig:real_testbeds}
\end{figure}

As shown in Figure \ref{fig:real_testbeds}, given an RGB image of the workspace, a gasket of fixed length $l_g$ and circular cross sectional diameter $d$, and a channel of width $w$, length $l_c$, and depth $h$ which is equal to at least $d$, we attempt to insert the gasket into the channel such that the gasket is completely contained within the channel. We assume that $l_g \approx l_c$ and $w \leq d \leq w + \delta$, where $\delta$ is a deformation constant determined by the cross-sectional compressability of the gasket. 

We denote the endpoints of the gasket as $g_0$ and $g_1$, and select two points in the channel, which we denote as $c_0$ and $c_1$. When the channel is open-ended (see Section \ref{subsec:channel} and Figure~\ref{fig:real_testbeds} A,B)  $g_0$ and $g_1$ correspond to the two endpoints of the channel; however, when the channel is closed (Figure~\ref{fig:real_testbeds} C), $g_0$ and $g_1$ correspond instead to adjacent points in the channel such that the inserted gasket, taking the \textit{longer} path between the two points, creates a closed loop.

We assume a planar work surface of known dimension, a six-axis robot arm with a parallel jaw gripper, and 3 RGB cameras with known intrinsic and extrinsic calibration matrices affixed at 1) the wrist of the robot, 2) above, and 3) to the side of the workspace. For the learned policy in addition to the 3 RGB cameras, we assume a human teleoperation input system such as a VR (Meta Quest2) controller or a 3D Mouse (SpaceMouse) or GELLO \cite{wu2023gello}.
We additionally constrain the problem space by assuming access to a 
priori knowledge of the shape and dimension of given channels, a non-adversarial (no sharp corners, knots, or crossings) gasket starting configuration located within the reachable workspace of the robot and the camera's field of view, and that the gasket and the channel can be easily color segmented from the workspace.

\section{Four Baseline Methods}
\label{sec:method}
\subsection{Learning Diffusion Policy}
Deep imitation learning where an agent learns required skills using deep neural networks trained on human demonstrations are being considered for a range of manipulation tasks including grasping, pressing, and pick and place~\cite{zhang_deep}. Various approaches have been used for deep imitation learning including Behavior Cloning, Generative Adversarial Imitation Learning (GAIL), Adversarial Reward-moment Imitation
Learning (AdRIL) and recently, diffusion policies\cite{arulkumaran2023pragmatic, correia2023survey, chi2023diffusionpolicy}, used in this paper.

The learned policy uses two ZED 2 stereo depth cameras and a Logitech BRIO webcam. One ZED 2 is mounted above the workspace and the other ZED 2 is mounted to the left of the workspace, with the UR5 and workspace fully in view. The Logitech webcam is mounted on the wrist of the UR5. A human operator teleoperates the robot using a 3D mouse. We use the Gello codebase ~\cite{wu2023gello} for teleoperation. 

A team of co-authors patiently perform the task 250 times using teleoperation and record all images and joint angles for training. This required approximately 15 hours of human effort. Human demonstrations are collected as described on the website.
The strategy used for the human demonstrations is closest to the Binary+ algorithm described in \ref{subsubsec:strategies}. As shown in Section \ref{subsec:results}, Binary+ is the best-performing of the three algorithms. We did this so we could properly compare the learned policy with the algorithms. Additionally, we focus on just the straight channel which is the simplest channel and collect all 250 demonstrations for that channel.

We use a CNN-based model architecture for Diffusion Policy with an action prediction horizon of 16 and observation history length of 2. $\mathbf{\epsilon}_\theta(.)$ takes in the noisy action $\mathbf{a}^k_{t, t+1, ..., t+15}$ and uses 36 1D convolutional layers with FiLM conditioning~\cite{perez2018film} on the observation embeddings ~\cite{chi2023diffusionpolicy}. For training, we set $K=100$ and use an initial learning rate of 0.0001 with a decay factor of 0.1 and we measure validation loss using Mean Squared Error(MSE). During execution, we use receding-horizon control with horizon 8. We use a Denoising Diffusion Implicit Model (DDIM)~\cite{song2020denoising} with 10 inference steps. These hyperparameter choices are from the default values as provided in~\cite{noauthor_robomimicrobomimicconfigdiffusion_policy_configpy_nodate}. The result is a policy that takes as input the camera images and generates as output control actions and gripper position. 
\subsection{3 Procedural Algorithms}

\begin{figure}
    \includegraphics[scale=0.28]{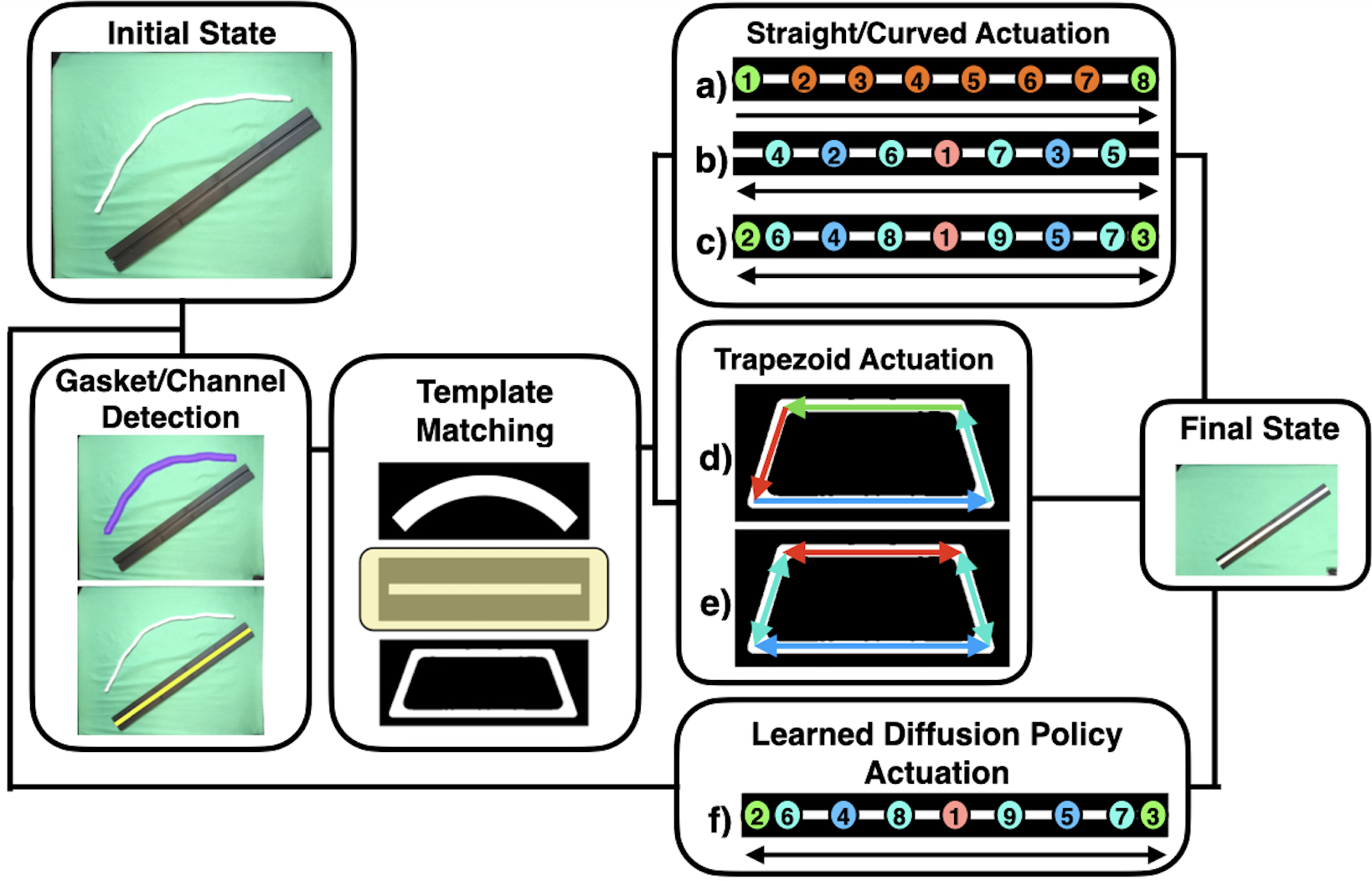}
    \caption{\textbf{Methods.} The Gasket/Channel Detection box shows gasket segmentation (above) and channel segmentation (below). The Template Matching box shows the three templates for the curved, straight and trapezoid channel. The Straight/Curved Actuation box shows selection and actuatio n strategies for the straight and curved channels: (a) is Unidirectional insertion, (b) is Binary search insertion, and (c) is Binary+ insertion. The colors on the channels represent the locations the robot attempts to place and press the gasket into while the numbers represent the order they are placed and pressed. Endpoints are green, midpoints are pink, half-points are blue and the quartile-points are cyan.  The arrows indicate the direction(s) of the slide(s). 
    For the trapezoid channel, we treat each segment of the trapezoid as an instance of the straight channel. In the unidirectional approach (d) we process each segment in a counterclockwise manner, starting at the blue segment. For hybrid and binary (e), we evaluate the blue segment, then the cyan segments, and finally the red segment. The learned policy proceeds directly from the initial state to actuation (f). The Final State box shows the final assembled gasket.} 
    \vspace{-0.7cm}
\label{fig:promethod}
\end{figure}

\subsubsection{Perception}
The perception system aims to detect, separate, and localize the channel and gasket given an RGB image from an overhead workspace camera. Since the channel shape is fixed, we use a template matching algorithm for channel localization and identification. To enable this, we generate a ground truth binary mask and aspect ratio for each channel by using the CAD source files for each part. These ground truth masks are passed to the perception pipeline and used in the classification and planning steps. We detail the approach:

\begin{enumerate}[label=(\roman*)]
    \item \textbf{Segmentation and Classification:} Given an RGB image of the scene, we apply a Gaussian smoothing operator with a $5\times 5$ kernel to reduce high-frequency sensor noise. Then, we threshold the image to have pixel values within $[100,255]$ to minimize the impact of the background color and apply a Canny edge detector. With the mask of edges in the scene, we use the Ramer–Douglas–Peucker algorithm to get contours for both the gasket and the mold containing the channel. Given these contours, we extract the smallest rectangle that encloses the input 2D point set, which can be treated as a detection bounding box defined by the box's center, dimensions, and rotation angle with respect to the x-axis. We find the aspect ratio of each detected bounding box and compare it to those of the given templates to classify the channel.
    \item \textbf{Localization via Alignment:} Given the detection bounding box information and inferred channel, the template is rescaled to match the dimensions of the observed bounding box, translated to align its center to that of the bounding box, and rotated to match the bounding box's rotation with respect to the x-axis. Once the template mask and channel are aligned, we can precisely localize the points in pixel space where the insertion channel exists in the physical channel. 
    \item \textbf{Skeletonization:} We skeletonize the localized channel to obtain a one-pixel-wide skeleton along the center of the channel. We then perform depth-first-search starting from an arbitrary pixel along this skeleton to get the pixels corresponding to the endpoints of the channel and order all points in the channel relative to these endpoints. This ordered list is necessary to relate pixels to their positions on the channel.
    \item \textbf{Waypoint Selection:} Using the ordered gasket skeleton, we sample a set of points between the two endpoints according to one of the insertion strategies (enumerated in section~\ref{subsubsec:strategies}). These points are used for the pick-place-press actions. 
\end{enumerate}

This perception pipeline runs after every step to provide an up-to-date estimate of the workspace state.

\subsubsection{Robot Primitives}
We define a small set of versatile primitives to enable efficient environment interaction.
\begin{enumerate}[label=(\roman*)]
    \item Pick and Place: The robot moves just above a point with grippers open, then descends to the height of the work surface and closes the grippers. The robot then rises, moves to a target point and opens the grippers.
    \item Shift and Place: In this variant of pick and pace, the robot slides along the surface until it reaches the target point to avoid lifting any adjacent gasket insertions.
    \item Press: The robot moves downward from above a target pose until it experiences an upward force above a given threshold (in our case, 40~N as measured by the robot's torque sensors).
    \item Slide: The robot moves sideways toward a target point across the work surface, while also pressing downward with a small amount of force (20~N).
    \item Home: The robot returns to a position above the workspace to afford an unobstructed view of the surface.
\end{enumerate}

\subsubsection{3 Insertion Algorithms}
\label{subsubsec:strategies} 
\textbf{Unidirectional} insertion, \textbf{binary search} insertion, and \textbf{Binary+}. These methods differ in the points selected for alignment, as well as the order in which the robot actuates to those points. (See Figure \ref{fig:promethod})

\begin{enumerate}[label=(\roman*)]
    \item Unidirectional insertion (Fig.~\ref{fig:promethod}a): The robot picks, places, and inserts the gasket into the channel, starting at one end of the gasket and progressing toward the other end. Then, the robot presses each selected point of the gasket into the channel for a second time to reinforce the insertion. Finally, the robot slides its gripper along the entire length of the channel to seal the gasket.
    
    \item Binary search insertion (Fig.~\ref{fig:promethod}b):
    The robot begins by picking and placing the midpoint of the gasket, followed by the points located at $\frac{1}{4}$ and $\frac{3}{4}$ of the gasket length, followed by those at the eighths, and so on until the algorithm reaches a termination limit (we set $\frac{1}{8}$ for the straight and curved channels, and $\frac{1}{4}$ for the trapezoid channel). Then, the robot presses each selected point of the gasket into the channel in the same order as they were picked and placed to reinforce the insertion. Lastly, the robot performs a ``binary slide'' by starting at the midpoint and sliding once toward each end.

    \item Binary+ insertion (Fig.~\ref{fig:promethod}c):
    This method attempts to combine the advantages of the unidirectional and binary search approaches. The robot begins by picking, placing, and inserting the midpoint of the gasket, followed by the endpoints. The robot then places the quarter and eighth points, as in the binary method. Then, the second reinforcing presses follow the unidirectional method. Finally, the binary slide is performed.

\end{enumerate}

\section{Physical Experiments}
\label{sec:experiment}

\subsection{Workspace Setup}
    We utilize a Universal Robots UR5 to conduct experiments. The UR5 is a 6-degree of freedom manipulator which can be operated in either position- or force-control modes to enable pressing and sliding motions. The work surface is aligned with the XY-plane and covered with a green tablecloth to make it perceptually uniform. The workspace is surrounded by a series of black drapes to control lighting and eliminate distractors. We found that the drapes were invaluable to the template matching. Additionally, we mount a front camera level with the workspace on the far side of the table from the robot. This camera is only used for evaluating insertion. A red backdrop is also used to increase contrast between the gasket and channel when viewed from the front camera for evaluation, and is not present during execution. 

The workspace is observed by three cameras during training and execution. The overhead camera is placed 97.1~cm above the center of the workspace and pointed downward so that it is able to observe the entire workspace. This is the only camera used during the execution of the procedural trials. The ``side'' camera is placed 40.3~cm outside the workspace on the left side and 32.7~cm above it, positioned such that the camera is able to see the entire workspace along with the robot. The ``wrist'' camera is mounted 16.0~cm above the end-effector on a fixed ``handle'' such that it observes the movement of the gripper along with the section of the workspace near the robot end-effector.

\subsection{Channels and Gasket}
\label{subsec:channel}
We consider 3 channels in increasing order of difficulty, as shown in Fig. \ref{fig:real_testbeds}. Fig. \ref{fig:real_testbeds}A: The first channel is an open straight channel with dimensions 26.5" x 2.68" x 0.56". Fig. \ref{fig:real_testbeds}B: The second channel is an open curved strut channel covering a $90^\circ$ arc of a circle, with inner diameter 32.4", outer diameter 35.1", and height 0.75". This results in a channel with an arc length of 26.5", a width of 2.68", and a height of 0.75", which is analogous to the dimensions of the straight channel. Fig. \ref{fig:real_testbeds}C: The third channel is a closed trapezoidal channel with a long side of 10", a short side of 7.5", and two 4.5"-long non-parallel sides. All channels have an inner channel width of 0.5". All channels are 3D printed from Black PLA using a Bambu Lab P1S FDM 3D printer. We use a white 0.5" braided nylon rope as our gasket analogue as it is deformable enough to meet the gasket-channel constraints as discussed in Section \ref{sec:cfp}. The rope is cut to a length of 26.5" to precisely match the length of the testbed channels. Figure \ref{fig:real_testbeds} shows the channels and gasket in the goal condition.

\subsection{100 Gasket Assembly Trials}
\label{subsec:exp_setup}
We only trained the demonstrations on the straight channel in a fixed position with various gasket initializations both above and below the channel. Thus, during testing, we only had the channel in the same fixed position. For Trials 1-10, i.e., for the learned policy, the channel is at a fixed pose of $0^\circ$. For Trials 11-100, the position and orientation of the channel are randomized at the beginning of each trial to any location completely within the reachable workspace and any angle within $\theta=\pm45^\circ$ of the horizontal, respectively.

At the beginning of all 100 trials, the starting position of the gasket is randomized. We perform this randomization by lifting the gasket with one fist and dropping it over either the top or bottom half of the workspace. The ends of the gasket are then moved outwards until the configuration of the gasket conforms to all of the constraints specified in Section~\ref{sec:cfp}. A trial ends after the robot successfully completes the task, the program terminates, or ten minutes has elapsed, whichever comes first.
For the unidirectional insertion, in the physical experiments, we always pick the left endpoint as a starting point but the algorithm can start from either endpoint. 

\begin{table*}[ht!]
  \centering
  \begin{tabular}{|c|c|c|c|c|c|c|c|c|c|c|}
  \hline \textbf{Trial No.} & 
  \textbf{Channel} & \textbf{Method} & \multicolumn{4}{c|}{\textbf{Alignment Performance}} & \multicolumn{4}{c|}{\textbf{Insertion Performance}}\\
  \hline
   & & & \textbf{0-25\%}& \textbf{25-50\%}& \textbf{50-75\%}\ & \textbf{75-100\%} & \textbf{0-25\%}& \textbf{25-50\%}& \textbf{50-75\%}\ & \textbf{75-100\%}\\
  \hline
  1-10 & Straight & Diffusion Policy &2&0&0&8 &2&0&1&7\\
  \hline
  11-20 & Straight & Unidirectional &6&0&1&3 &6&1&1&2\\
  21-30 & Straight & \textbf{Binary Search} &0&0&0&\textbf{10} &0&0&0&\textbf{10}\\
  31-40 & Straight & \textbf{Binary+} &0&0&0&\textbf{10} &0&0&0&\textbf{10}\\
  \hline
  41-50 & Curved & Unidirectional &5&1&0&4 &7&0&1&2\\
  51-60 & Curved & Binary Search &0&0&4&6 &1&3&6&0\\
  61-70 & Curved & \textbf{Binary+} &0&0&1&\textbf{9} &0&0&1&\textbf{9}\\
  \hline
  71-80 & Trapezoid & Unidirectional &10&0&0&0 &10&0&0&0\\
  81-90 & Trapezoid & Binary Search &9&1&0&0 &9&\textbf{1}&0&0\\
  91-100 & Trapezoid & Binary+ &9&0&\textbf{1}&0 &9&\textbf{1}&0&0\\
 \hline
  \end{tabular}
  \caption{\textbf{Alignment and Insertion Performance of 100 Physical Trials.}}
  \label{tab:tab1}
  \vspace*{-4pt}
\end{table*}

Lightproof curtains on all sides of the robot are kept closed during experiments, with the exception of the curtains directly in front of the robot, which are kept open so as not to obstruct the view of the ``front'' workspace camera and to facilitate the placement of a workspace key light. During data collection for the learned policy, we additionally open a small section of drapes behind the ``side'' camera, from which the human demonstrator can view the workspace for teleoperation. We maintain this during evaluation to ensure that the environment for running trials was as similar as possible to the environment in which the human demonstrations were collected for best test time results. %

\subsection{Experimental Evaluation Metrics}
\label{subsec:metric}
Since automated evaluation methods were prone to error, we developed two manually-evaluated performance metrics.

After the robot execution has terminated, a human judge visually rates performance into one of four alignments categories and one of four insertion categories.

We note that given the specifications of \TaskName{} there is a relationship between alignment and insertion. In order for the gasket to be properly inserted, it must first have been well aligned. The quality of alignment affects the resultant quality of insertion.

\section{Results}
\label{subsec:results}
\subsection{Alignment and Insertion}
\label{subsec:alignment+insertion}
We perform 100 physical trials: 10 for the learned diffusion policy on the straight channel in fixed $0^\circ$ pose and 90 across all procedural approaches and all channel types with varying channel pose positions and orientations as noted in Section \ref{subsec:exp_setup}. See results in Table~\ref{tab:tab1}. We again note that the learned policy was only evaluated for a fixed channel pose which matched the pose used during data collection whereas for the procedural algorithms, the channel pose varied significantly.

For the straight channel, the binary search  and Binary+ approaches achieve 75-100\% in all trials. For the curved channel, the Binary+ approach separates itself from the binary search approach, attaining the highest alignment and insertion performance. Finally, for the most difficult channel, the trapezoid, the Binary+ approach attains the best alignment performance, while having the same outcomes as the binary search for the insertion performance.

For the diffusion policy (Trials 1-10), there were three failed trials across the two metrics. These failures occurred because the diffusion policy failed to fully insert the gasket within the 10-minute threshold. In all three instances, the failure stemmed from incorrect execution of the pick-place-press sequence. Specifically, the robot either did not pick up the gasket, picked it but failed to place it correctly in the channel, or pressed in the wrong position, disrupting the subsequent pick-place-press sequences. Additionally, the robot could not recover from these failures, often spending several minutes attempting to improve the press without success or failing to detect the error.

Similarly, for the procedural methods on the straight and curved channels (Trials 11-70), the failures stemmed from failing to press the gasket into the channel at some points along the channel, leaving raised portions. When executing slide primitive in these scenarios, the raised portions caught on the gripper, causing large portions of the gasket to be unseated from the channel. For the trapezoid channel (Trials 71-100), both the binary search and Binary+ methods were initially able to insert the gasket into the long side. However, when the robot would try to round the corners to go to the other sides of the trapezoid, it inadvertently unseated the already inserted parts of the gasket. These errors compounded across all four sides of the trapezoid channel. 

\subsection{Completion Time Analysis}
\label{subsec:time_analysis}
We observe a difference in the time required to complete the task when comparing the learned and procedural methods. The learned diffusion policy ran until the task was completed or it was terminated when the maximum time horizon of 10 minutes was reached. Across all 10 trials, the average completion time is approximately 5 minutes 34 seconds. This includes the three trials that terminated at the maximum time horizon (10 minutes) without task completion; excluding these trials reduces the average completion time to 3 minutes 40 seconds. The procedural algorithms ran for approximately 3 minutes 30 seconds for the straight and curved channels and 7 minutes for the trapezoid.

\subsection{Discussion}
\label{subsec:discussion}
For the learned diffusion policy, four humans provided 250 demonstrations. 

With the unidirectional algorithm, we observe that after the first pick and place of an endpoint, most of the gasket still remains on the table. Consequently, the first press into the channel often fails to seat the gasket properly. This misalignment makes the next part of the gasket difficult to pick since it is positioned right alongside the channel. Moreover, even if the robot does grab the next point, since the previous pick-place-press does not properly seat the gasket, moving the next point along the gasket for pick and place can cause that previous segment to be completely unaligned from the channel. Since the unidirectional policy never revisits this prior point for pick and place, the push will very likely fail as well. Alternatively, with binary search, since the robot picks the gasket first at the middle, a larger amount of the gasket is now on the channel—approximately double compared to the unidirectional policy's initial pick. This increases the likelihood that the midpoint will be properly inserted when pressed. This proper seating at the midpoint is crucial for successful assembly, particularly in straight and curved channels. Compared to binary search, the Binary+ policy has more pick-place-pushes at points close to but not exactly at points that the binary search already reaches. This corrects any errors that the binary search might leave. Additionally, the Binary+ policy intentionally carries out pick-place-press at the endpoints, which ensures that when the robot executes the final slide, the grippers do not get caught at an unseated endpoint and drag the entire gasket out of the channel.
\section{Limitations + Future Work}
In this paper, we present a new \TaskName{} problem and provide results from 4 methods, a learned diffusion policy and 3 procedural algorithms. 
In the future, we will perform more experiments varying channel pose for the diffusion policy and learning diffusion policies for the curved and trapezoidal channels. We also plan to better handle recovery from poorly executed primitives by exploring additional approaches including hierarchical imitation learning~\cite{luo2023multistage} and self-supervised learning. We will also explore more complex channel shapes and sizes, work on a perception system that is more robust against different lighting conditions and distractors and examine if various gripper types have any significant effect on the task of \TaskName{}.  
\section*{Acknowledgements}
\vspace{-0.1cm}
\relsize{-1}
This research was performed at the AUTOLAB at UC Berkeley in affiliation with the Berkeley AI Research (BAIR) Lab, and the CITRIS "People and Robots" (CPAR) Initiative. We thank our colleagues who provided helpful feedback, help and suggestions, in particular Lawrence Chen, Philip Wu, Kaushik Shivakumar,  Justin Kerr and Chung Min Kim.

\renewcommand*{\bibfont}{\footnotesize}
\printbibliography
\clearpage
\normalsize
\section{Appendix}
\label{sec:appendix}
\subsection{Human Demonstrations}
Human demonstrations are collected as follows:
\begin{enumerate}
    \item The channel is fixed in place horizontally across the workspace, separating the workspace into a lower and upper section.
    \item The gasket is randomly dropped in either the lower or upper section so that it does not overlap itself and does not touch the channel. 
    \item The midpoint of the gasket is grasped and placed on top of the midpoint of the channel. The gripper then presses the gasket down into the channel.
    \item One endpoint of the channel is chosen arbitrarily. The gasket is placed on top of the selected endpoint of the channel and pressed down into the channel.
    \item The remaining endpoint of the gasket is then placed on top of the other endpoint of the channel and pressed to insert it into the channel.
    \item The gripper is moved to the quartile points (the order of the quartile points the gripper goes to is chosen arbitrarily) and pressed down on the gasket such that at those points the gasket is inserted into the channel.
    \item The gripper goes to the 'eighth' points (again the order of the points the gripper goes to is chosen arbitrarily) and presses down on the gasket such that at those points the gasket is inserted into the channel.
    \item The gripper goes to the midpoint of the gasket, moves down slowly to the channel surface such that the gripper touches the channel surface, and moves horizontally with no vertical movement towards one of the endpoints of the channel (chosen arbitrarily). The gripper returns to the midpoint of the channel and repeats this motion towards the other endpoint of the channel. This 8-step procedure is repeated for each human demonstration.
\end{enumerate}

\subsection{Experimental Evaluation Metrics Breakdown}
After the robot execution has terminated, a human judge visually rates performance into one of four alignments categories, as follows:

\begin{enumerate}
    \item \textbf{0\% - 25\%:} A \textbf{major alignment failure}, in which the robot has successfully aligned \textbf{less than 25\%} of the gasket with the channel.
    \item \textbf{25\% - 50\%:} A \textbf{partial alignment failure}, in which \textbf{between 25\% and 50\%} of the gasket has been successfully aligned.
    \item \textbf{50\% - 75\%:} A \textbf{partial alignment success}, in which \textbf{between 50\% and 75\%} of the gasket has been properly aligned.
    \item \textbf{75\% - 100\%:} A \textbf{full alignment success}, in which the robot has properly aligned \textbf{at least 75\%} of the gasket length with the channel.
\end{enumerate}

Similarly, a human judge visually rates performance into one of four insertion categories, as follows::
\begin{enumerate}
    \item \textbf{0\% - 25\%:} A \textbf{major insertion failure}, in which \textbf{less than 25\%} of the gasket is inserted into the channel.
    \item \textbf{25\% - 50\%:} A \textbf{partial insertion failure}, in which \textbf{between 25\% and 50\%} of the gasket is inserted.
    \item \textbf{50\% - 75\%:} A \textbf{partial insertion success}, in which \textbf{between 50\% and 75\%} of the gasket is inserted.
    \item \textbf{75\% - 100\%:} A \textbf{full insertion success}, in which \textbf{at least 75\%} of the gasket length is inserted.
\end{enumerate}

\begin{figure}
    \centering
    \includegraphics[scale=0.15]{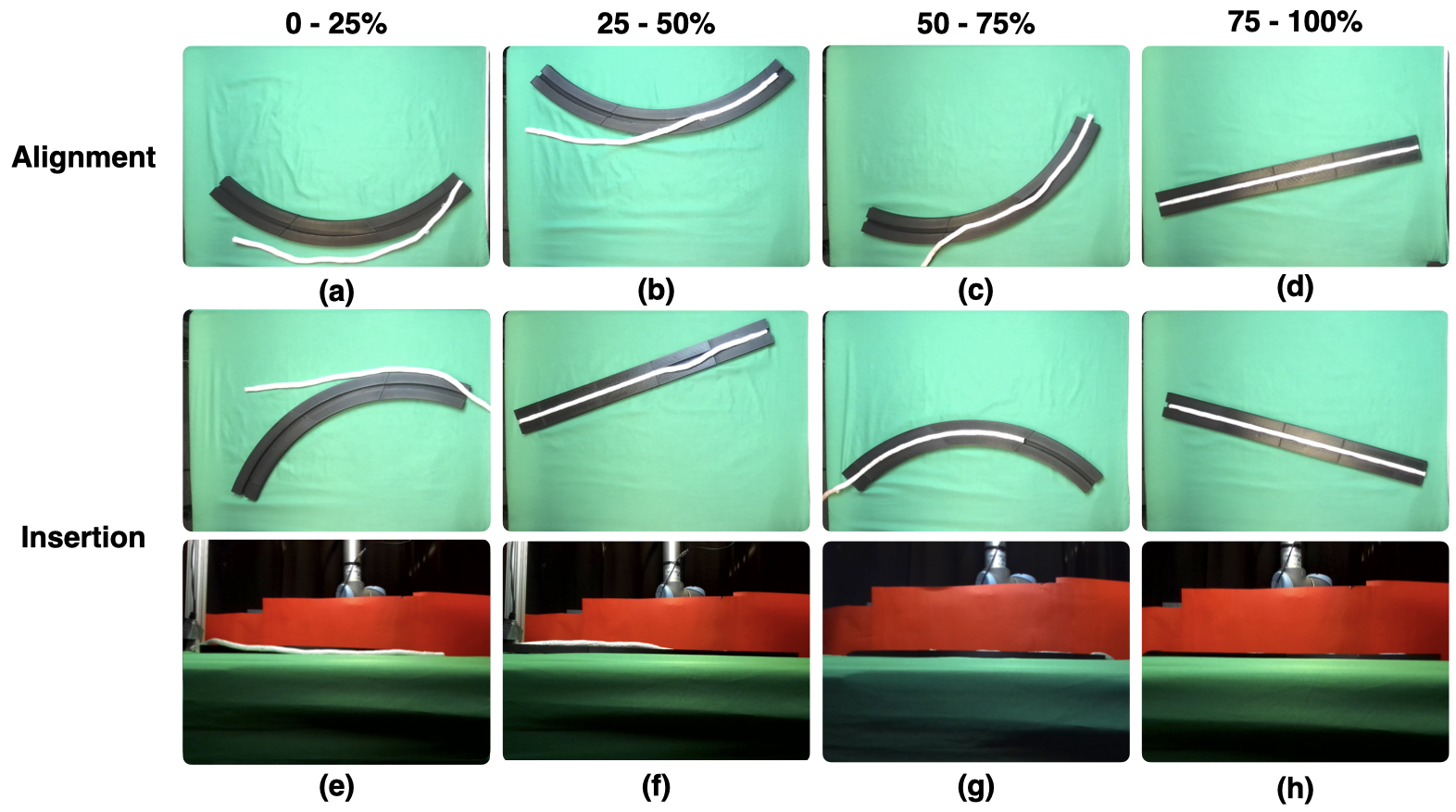}
    \caption{\textbf{Evaluation Metric Examples.} We provide examples for all four categories of the alignment and insertion evaluation metrics discussed in Section \ref{subsec:metric}. We show the final gasket and channel states after the robot attempts gasket assembly. For alignment we only consider the view from the overhead camera to determine alignment between the gasket and channel. To determine the snug fit of the insertion, we consult both the overhead view (top row) and the front view (bottom row), because (f), for example, shows how a gasket that is aligned with the channel can have poor insertion.}
    \label{fig:evaluation_metric}
    \vspace{-0.55cm}
\end{figure}

Figure \ref{fig:evaluation_metric} shows qualitative results from the trials of the three analytical algorithms in increasing order of success.
\end{document}